\documentclass[conference]{IEEEtran}
\IEEEoverridecommandlockouts

\usepackage{cite}
\usepackage{amsmath,amssymb,amsfonts}
\usepackage{algorithm}
\usepackage{algpseudocode}
\usepackage{graphicx}
\usepackage{textcomp}
\usepackage{xcolor}
\usepackage{booktabs}
\usepackage{multirow}
\usepackage{makecell}
\usepackage{url}
\graphicspath{{../../}{./}}
\def\BibTeX{{\rm B\kern-.05em{\sc i\kern-.025em b}\kern-.08em
    T\kern-.1667em\lower.7ex\hbox{E}\kern-.125emX}}
\begin{document}
\bstctlcite{IEEEexample:BSTcontrol}

\title{Delta2Gamma: Band-Wise Adaptive Contrastive Learning of EEG for Alzheimer's Disease Detection}

\author{\IEEEauthorblockN{Chanwoo Park}
\IEEEauthorblockA{\textit{Department of Artificial Intelligence} \\
\textit{Korea University}\\
Seoul, Republic of Korea \\
cksdn1290@korea.ac.kr}
\and
\IEEEauthorblockN{Chanwoo Kim}
\IEEEauthorblockA{\textit{Department of Artificial Intelligence} \\
\textit{Korea University}\\
Seoul, Republic of Korea \\
chanwcom@korea.ac.kr}
}

\maketitle

\begin{abstract}
Low-cost, scalable screening for dementia remains an open problem.
Imaging-based diagnosis is costly and hard to deploy widely.
Electroencephalography (EEG) is portable and inexpensive, but its recordings are noisy, vary widely across subjects, and carry few clinical labels.
We tackle this with Delta2Gamma, a self-supervised framework that learns EEG representations from unlabeled data by contrasting augmented views of each signal.
Rather than treat EEG as a single stream, Delta2Gamma decomposes every recording into the five canonical neural rhythms (delta, theta, alpha, beta, gamma).
Each band gets its own encoder and projection head.
Each also gets a temperature that is predicted adaptively during contrastive training, so bands with different signal statistics are balanced automatically.
On the ADFTD cohort under a strict leave-one-subject-out protocol, Delta2Gamma separates Alzheimer's disease from cognitively normal controls with 92.4\% accuracy.
This exceeds both supervised backbones and recent dedicated EEG methods.
\end{abstract}


\section{Introduction}
Global aging and the explosive growth in dementia prevalence form a public-health crisis that is straining socioeconomic structures and healthcare systems worldwide \cite{nichols2022estimation, li2022global}. Traditional diagnostic methods such as magnetic resonance imaging (MRI) and positron emission tomography (PET) have limited value for early detection because of their high cost, limited accessibility, and dependence on large, stationary equipment and specialized personnel \cite{haidar2023asl}. Because these modalities are confined to specialized medical centers, they cannot serve as frontline screening tools for the broad at-risk population. As a result, early-stage mild cognitive impairment, the precursor to dementia, is difficult to catch, and most patients are diagnosed only after their symptoms have substantially progressed \cite{sabbagh2020early}.

Electroencephalography (EEG) offers a scalable alternative. It is non-invasive and directly and quantitatively measures declines in brain function \cite{dabbabi2023review}, unlike functional MRI, which captures blood-flow changes that only indirectly proxy neural activity. By directly recording the electrical signals generated by neurons, EEG reflects how the neuropathological changes that define dementia impair brain function, providing critical insight into the neurophysiological basis of cognitive decline \cite{smailovic2019neurophysiological}. These properties make EEG well suited to portable, repeatable, and affordable cognitive monitoring outside the hospital.
Other low-cost digital biomarkers have been pursued in parallel: large language models (LLMs) applied to spontaneous speech detect AD with chain-of-thought reasoning over transcribed narratives \cite{park2025cotspeech}, and pairing them with vision-language models for picture-description tasks improves detection further \cite{park2025cotvlm}; clinically inspired cognitive test batteries have even been repurposed to evaluate the reasoning capacity of the LLMs themselves \cite{park2027lmmse}. These behavioral markers, however, rely on language production and task compliance, whereas EEG measures the underlying neural dysfunction directly, which motivates our focus on resting-state recordings.

A large body of work identifies quantifiable spectral signatures of Alzheimer's disease (AD) and other dementias, summarized as an overall \emph{slowing} of brain oscillations. Patients show increased power in low-frequency bands, delta ($\delta$, 0--4 Hz) and theta ($\theta$, 4--8 Hz) \cite{moretti2004individual}, together with reduced power in higher-frequency bands, including alpha ($\alpha$, 8--12 Hz), beta ($\beta$, 12--30 Hz), and a marked reduction in gamma ($\gamma$, $>$30 Hz) that is tightly linked to higher cognitive functions such as short-term memory \cite{traikapi2021gamma}. These band-specific changes motivate a model that treats each rhythm separately rather than collapsing the signal into a single representation. Motivated by these biomarkers, the main contributions of this paper are:
\begin{enumerate}
    \item \textbf{Frequency-band-specific encoding}: we propose the Delta2Gamma architecture, which decomposes EEG into the five canonical bands ($\delta, \theta, \alpha, \beta, \gamma$) and processes each with an independent CNN encoder and projection head, preserving band-specific neural information.
    \item \textbf{Effective dementia classification}: we design a self-supervised model that captures the neurophysiological hallmarks of dementia (increased $\delta/\theta$, decreased $\alpha/\beta$), achieving state-of-the-art accuracy under strict subject-independent evaluation.
\end{enumerate}

\section{Proposed Method}
\begin{figure}[t]
    \centering
    \includegraphics[width=0.93\linewidth]{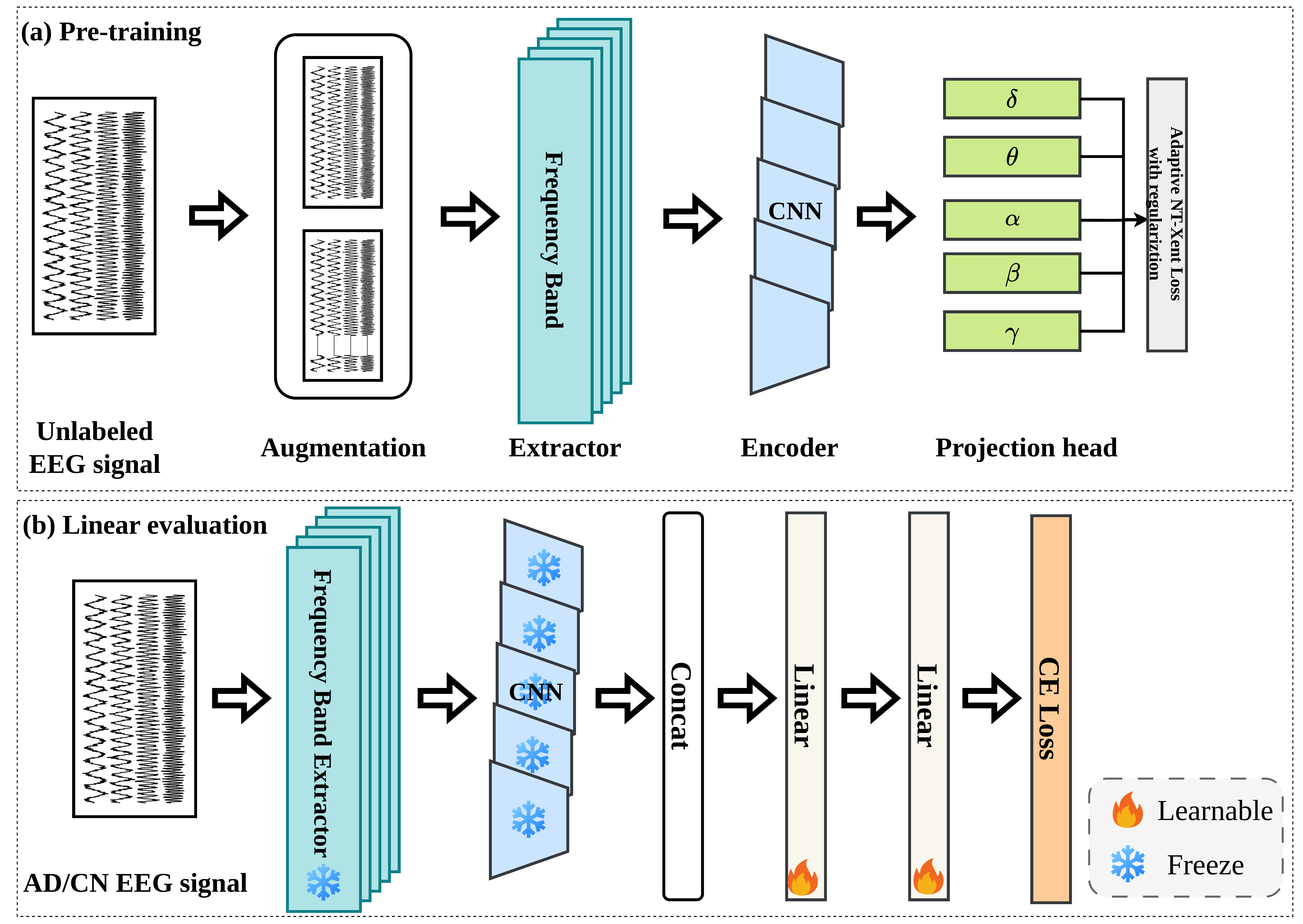}
    \caption{Overview of the Delta2Gamma model. (a) Self-supervised pre-training: augmentation is applied to unlabeled EEG, then an adaptive NT-Xent contrastive loss with regularization trains the encoder to learn representations across the five bands ($\delta,\theta,\alpha,\beta,\gamma$). (b) Linear evaluation: a classifier is added on top of the frozen pre-trained encoder for AD vs. cognitively normal (CN) classification. Numbers on linear layers denote layer dimensionality.}
    \label{fig:1}
\end{figure}

The framework (Fig.~\ref{fig:1}) has two stages: self-supervised pre-training and linear evaluation. We perform contrastive learning on unlabeled EEG so the model learns general signal characteristics, while data augmentation yields representations robust to transformations.

\subsection{Self-Supervised Multi-Band Architecture}
We build on SimCLR \cite{chen2020simple}, a contrastive self-supervised method that pulls augmented views of the same sample together while pushing different instances apart. Our adaptation (Fig.~\ref{fig:1}(a)) is tailored to EEG: the model simultaneously processes the five frequency bands and uses an independent projection head per band for fine-grained feature learning. This is especially valuable in clinical settings, where unlabeled EEG is abundant but labels are scarce. The pre-training procedure is given in Algorithm~\ref{alg:ssl_pretraining}.

A raw multi-channel EEG signal of shape $[C,L]$ ($C{=}19$ channels; $L$ = seconds $\times$ 500\,Hz sampling rate) is decomposed into five canonical bands ($\delta$: 0.5--4, $\theta$: 4--8, $\alpha$: 8--13, $\beta$: 13--30, $\gamma$: 30--45 Hz) by bandpass filters, giving five parallel views of shape $[5,C,L]$. The band extractor uses parallel 1-D depthwise convolutions (kernel size 7, padding 3, groups $=C$) followed by batch normalization \cite{ioffe2015batch} and ReLU \cite{agarap2018deep}, so each channel is processed independently to preserve its temporal patterns \cite{Lawhern2018EEGNet}. Each band is then passed through an independent encoder of three convolutional blocks with increasing channels ($32{\rightarrow}64{\rightarrow}128$), interleaved with batch normalization, ReLU, and max pooling, producing $[5,C,L/32]$. Global average pooling summarizes temporal dynamics, and the five pooled outputs are fused by a fully connected layer with batch normalization and ReLU into a compact $5\times128$-dimensional embedding, one projection vector per band.

\begin{algorithm}[t]
\caption{SimCLR-based Self-Supervised Pre-training}
\label{alg:ssl_pretraining}
\begin{algorithmic}[1]
\Require Unlabeled EEG $\mathcal{D}$, batch size $B$, epochs $E$
\Ensure Pre-trained multi-frequency encoder $f_{\theta}$
\State Init encoder $f_{\theta}$ with bands $\{\delta,\theta,\alpha,\beta,\gamma\}$, heads $\{g_k\}_{k=1}^{5}$, temperature nets $\{\phi_k\}_{k=1}^{5}$
\For{$e=1$ to $E$}
  \For{each mini-batch $\mathcal{B}=\{x_i\}_{i=1}^{B}$, $x_i\in\mathbb{R}^{C\times T}$}
    \State $\mathcal{B}^{(1)}\!\leftarrow$ weak-aug$(\mathcal{B})$;\; $\mathcal{B}^{(2)}\!\leftarrow$ strong-aug$(\mathcal{B})$
    \For{view $v\in\{1,2\}$, band $b$}
        \State $\mathbf{x}_b^{(v)}\!\leftarrow$ BandpassFilter$_b(\mathcal{B}^{(v)})$
        \State $\mathbf{z}_b^{(v)}\!\leftarrow g_b(\text{CNN}_b(\mathbf{x}_b^{(v)}))$
    \EndFor
    \For{each band $b$}
        \State $\tau_b\leftarrow\phi_b(\mathbf{z}_b^{(1)},\mathbf{z}_b^{(2)})$
        \State $\mathcal{L}^{(b)}\!\leftarrow$ NTXent$(\mathbf{z}_b^{(1)},\mathbf{z}_b^{(2)},\tau_b)+\lambda\big(\tfrac{d}{2}\log\tau_b+\tfrac{1}{\tau_b}\big)$
    \EndFor
    \State Update parameters w.r.t. $\mathcal{L}_{\text{total}}=\sum_b \mathcal{L}^{(b)}$
  \EndFor
\EndFor
\State \Return $f_{\theta}$
\end{algorithmic}
\end{algorithm}

\subsection{Data Augmentation}
The core of self-supervised learning is to extract a meaningful signal from the data itself through a pretext task. For each EEG signal we generate two semantically identical but morphologically different views using Gaussian noise (std 0.03), amplitude scaling (factor in $[0.8,1.2]$), and 10\% random masking in both the time and frequency domains; with 10\% probability we also drop 10\% of the channels. The two views form an instance-discrimination task: views of the same signal are positive pairs and views of different signals are negatives, driving the model toward noise- and transformation-robust representations. By contrasting a weak and a strong view of each signal, the encoder is encouraged to discard nuisance variation while preserving the band-specific structure relevant to diagnosis.

\subsection{Training Objective}
The pre-training loss aggregates per-band adaptive NT-Xent losses with temperature regularization \cite{wang2024adaptive}: $\mathcal{L}_{\text{pt}}=\sum_{b=1}^{B}\mathcal{L}_b$, where
\begin{align}
\mathcal{L}_{\text{pt}} = \sum_{b=1}^B \Big(
    & -\tfrac{1}{\tau^{b+}} \, \mathrm{sim}(\mathbf{z}^{b}, \mathbf{z}^{b+}) \notag \\
    & + \tfrac{1}{\tau^{b-}_{n^*}} \max_{n=1,\cdots,N} \mathrm{sim}(\mathbf{z}^{b}, \mathbf{z}^{b-}_{n}) \notag \\
    & + \beta \, \Omega(\tau^{b+}) - \beta \, \Omega(\tau^{b-}_{n^*})
\Big),
\label{eq:pre-train_loss}
\end{align}
$\mathrm{sim}(\cdot,\cdot)$ is cosine similarity over positive pair $(\mathbf{z}^{b}, \mathbf{z}^{b+})$ or negative pair $(\mathbf{z}^{b}, \mathbf{z}^{b-}_{n})$; $N$ is the number of negatives; $\tau^{b+}$ and $\tau^{b-}_{n}$ are learnable adaptive positive and negative temperatures with $n^*=\arg\max_{n}\mathrm{sim}(\mathbf{z}^{b},\mathbf{z}^{b-}_{n})$; and $\beta\geq0$ controls the regularizer
\begin{equation}
\Omega(\tau)=(d'/2)\log(\tau)+1/\tau,
\end{equation}
where $d'$ is the projected dimension; this drives $\tau\rightarrow 2/d'$. Unlike fixed-temperature SimCLR, each band receives a dynamically predicted temperature reflecting its distribution and learning difficulty. Each band therefore receives a contrastive signal whose strength is matched to its own statistics, resolving the uniformity--tolerance trade-off that a single global temperature cannot address and stabilizing training across bands with very different power profiles.

\subsection{Downstream Task}
For classification we attach a three-layer MLP on top of the pre-trained encoder (hidden sizes 512 and 256, each with ReLU \cite{agarap2018deep}, batch normalization, and dropout \cite{srivastava2014dropout} of 0.3 and 0.2), with an output layer sized to the number of classes. We consider both freezing the encoder (linear evaluation, Fig.~\ref{fig:1}(b)) and fine-tuning all parameters.

\section{Experimental Setup}
Pre-training uses AdamW \cite{loshchilov2017decoupled} (batch size 64, learning rate $1\times10^{-4}$), the adaptive NT-Xent loss with temperature in $[0.05,0.5]$ and $\beta{=}0.01$, weight decay $1\times10^{-5}$, and cosine annealing with warm restarts. Linear evaluation uses Leave-One-Subject-Out (LOSO) cross-validation with frozen encoder weights, AdamW (batch size 32, learning rate $1\times10^{-4}$), and cross-entropy loss. Both stages run up to 100 epochs with early stopping (patience 10) on an NVIDIA RTX 4090 GPU.

\subsection{Dataset}
We use the publicly available ADFTD dataset \cite{miltiadous2023dataset}: resting-state, eyes-closed recordings from 88 participants (36 AD, 23 frontotemporal dementia [FTD], 29 CN). Mean MMSE scores were 17.75 (AD), 22.17 (FTD), and 30 (CN). Signals were recorded with a Nihon Kohden EEG-2100 system using 19 scalp electrodes (10--20 system) at 500\,Hz with impedance below 5\,k$\Omega$. Because FTD is hard to identify from EEG alone, classification focuses on AD vs. CN; the 23 FTD subjects (not used for classification) provide unlabeled data for self-supervised pre-training.

\subsection{Preprocessing and Segmentation}
We compute an average reference across all channels, since referencing strongly affects amplitude measurements, and apply a 6th-order Butterworth bandpass filter (0.5--45 Hz) \cite{nour2024novel} that preserves the activity relevant to distinguishing AD from CN. Ocular and muscular artifacts are then removed via blind source separation with independent component analysis \cite{ica1}, all within MNE-Python \cite{GramfortEtAl2013a}.
Eyes-closed EEG is segmented into 30-second epochs, matching the standard epoch length used in sleep research \cite{park2025sleep} and minimizing the influence of external stimuli \cite{ye2023dementia}.

\subsection{Evaluation Protocol and Metrics}
LOSO cross-validation \cite{Kunjan2021} trains on $N{-}1$ subjects and tests on the held-out subject, repeating for all subjects. By preventing cross-subject data leakage, it provides a stringent, realistic estimate of generalization under high inter-individual variability. We report accuracy, precision, recall, weighted F1-score, and AUC.

\section{Results and Discussion}
\subsection{LOSO Performance}
We compare against major EEG benchmarks implemented with Braindecode \cite{Schirrmeister2017Deep} and prior AD-vs-CN studies, fine-tuning self-supervised baselines where pretrained weights are available.
As summarized in Table~\ref{tab:result1}, common supervised and self-supervised EEG backbones (e.g., ATCNet \cite{Altaheri2022Physics}, EEGNet \cite{Lawhern2018EEGNet}, EEGConformer \cite{Song2022EEG}, BIOT \cite{yang2023biot}, Labram \cite{jiang2024large}, S-JEPA \cite{guetschel2024sjepa}) reach only 39--74\% accuracy on this task.
Under strict LOSO cross-validation, our adaptive 5-band-head model attains 92.37\% accuracy and 92.33\% F1-score, surpassing recent dedicated AD-vs-CN methods (Table~\ref{tab2}).
This shows the multi-band self-supervised approach is highly effective for the complex features of EEG and generalizes well despite inter-subject variability.

\begin{table*}[htbp!]
\centering
\begin{tabular}{lcccccc}
\toprule
\textbf{Model} & \textbf{Application} & \textbf{Train} & \textbf{Backbone} & \textbf{\#Param}& \textbf{Accuracy (\%)} & \textbf{F1-score (\%)} \\
\midrule
ATCNet \cite{Altaheri2022Physics}  & General & Supervised   & CNN \& RNN \& Att & 113,732 & 74    & 74   \\
BIOT \cite{yang2023biot} & \makecell{Sleep Staging \\ Epilepsy} & SSL & Att & 3,183,879 & 53   & 40\\
CTNet \cite{zhao2024ctnet} & Motor Imagery & Supervised & CNN \& Att & 26,900 & 74    & 73   \\
Deep4Net \cite{Schirrmeister2017Deep} &  General & Supervised & CNN   & 282,879          & 49    & 49   \\
EEGConformer \cite{Song2022EEG}       & General & Supervised & CNN \& Att & 789,572  & 57 & 54   \\
EEGInception \cite{santamariavazquez2020eeginception} & \makecell{Motor Imagery \\ ERP \& SSVEP} & Supervised & CNN   & 558,028     & 39 & 37   \\
EEGNet \cite{Lawhern2018EEGNet}  & General & Supervised & CNN    & 2,484   &46 & 45   \\
FBCNet \cite{mane2021fbcnet} & Motor Imagery & Supervised & CNN & 11,812 & 48 & 38 \\
Labram \cite{jiang2024large}   & General & SSL & CNN \& Att & 5,866,180 & 54 & 38  \\
S-JEPA \cite{guetschel2024sjepa}& \makecell{Motor Imagery \\ ERP \& SSVEP} & SSL  & CNN \& Att  & 3,456,882   & 50 & 50  \\
SPARCNet \cite{jing2023development} & Epilepsy & Supervised & CNN  & 1,141,921   & 54 & 53  \\
TCN \cite{Bai2018Empirical}   & General & Supervised & CNN \& RNN & 26,974 & 44 & 40  \\
TIDNet \cite{Kostas2020Thinker} & General & Supervised & CNN  & 240,404   & 44 & 40  \\
\midrule
\textbf{Ours (adaptive 5 band heads)} & Dementia & SSL & CNN & 976,635 & \textbf{92} & \textbf{92}  \\
\bottomrule
\end{tabular}
\caption{AD vs. CN classification performance of the proposed model against leading supervised and self-supervised EEG benchmarks. “Application" denotes the model's typical target domain, with “General" indicating no specific one; “\#Param" is the number of parameters required to instantiate the model. Att denotes attention mechanisms.}
\label{tab:result1}
\end{table*}

\begin{table}[htb!]
\centering
\caption{LOSO comparison for AD vs. CN classification on the ADFTD dataset \cite{miltiadous2023dataset}.}
\label{tab2}
\setlength{\tabcolsep}{4pt}
\begin{tabular}{lcccc}
\toprule
\textbf{Model} & \textbf{Acc (\%)} & \textbf{F1 (\%)} & \textbf{Prec. (\%)} & \textbf{Rec. (\%)} \\
\midrule
kNN \cite{ntetska2025complementary}              & 60.30    & 58.90 & 57.90 & 59.90   \\
CNN \cite{stefanou2025novel}              & 79.45    & 77.60 & 76.32 & 76.06   \\
Random Forest \cite{sarkar2025detection}            & 80.00    & 81.69 &   --&   80.55   \\
DICE-Net \cite{miltiadous2023dice}              & 83.28    & 84.12 & 88.94 &  79.81  \\
CNN \cite{vo2025multimodal}              & 84.62    & 86.11 & -- & 86.11   \\
MJANet \cite{sun2025enhanced}              & 85.23    & 86.37 & 88.12 &  84.69  \\
Dual-Branch \cite{chen2023multi}             & 85.78    &   -- &   --&   83.22 \\
Random Forest \cite{parihar2024analysis}            & 88.90    &   -- &   --&   --   \\
BI-MCGNN \cite{zhang2025dual}            & 91.25    &   -- &   --&   \textbf{93.32}   \\
\midrule
\textbf{Ours} & \textbf{92.37} & \textbf{92.33}  & \textbf{92.61} & 92.37\\
\bottomrule
\end{tabular}
\end{table}

\subsection{Multi-Band Feature Analysis}
The band-specific encoders learn distinct activation strategies. The $\delta$ encoder exhibits a restricted activation range concentrated on a few features, the $\theta$ band is more uniformly distributed, the $\alpha$ band shows the highest and most consistent activations, and the $\beta$ and $\gamma$ bands remain at intermediate but irregular levels, indicating that each band adopts an encoding scheme matched to its own signal characteristics. Group-wise importance analysis is more revealing: peak importance in the $\delta$ band ($\approx2\times10^{-4}$) is about twenty times that of the $\gamma$ band ($\approx1\times10^{-5}$), indicating that low-frequency bands carry stronger diagnostic cues for dementia. Importantly, highly activated features are not necessarily the most diagnostic, suggesting the model separates general signal processing from the extraction of pathological patterns. The learned feature-correlation structure echoes this: low-frequency bands ($\delta$, $\theta$) condense information into dense, near-diagonal correlations, whereas high-frequency bands ($\beta$, $\gamma$) form sparse, largely independent representations that capture complementary, non-redundant aspects of the signal. Topographically, the AD group shows localized $\delta$-band increases over frontal, temporal, and parietal regions, whereas CN $\delta$ activity is weaker and posteriorly distributed, consistent with known biomarkers. Together these observations give the multi-band design both neurobiological plausibility and clinical utility.

\subsection{Ablation Study}
Table~\ref{tab:adcn_comparison} isolates the contribution of each component. A CNN trained from scratch without self-supervision reaches only 62.54\% accuracy, whereas our full model reaches 92.37\%, a 47.70\% relative gain that confirms the value of contrastive pre-training when labels are scarce. Replacing the five band-specific heads with a single projection head drops accuracy to 89.65\% versus 91.14\% for five heads, and the adaptive model improves a further 3.03\% (relative) over the single-head baseline, showing that per-band heads capture complementary frequency information. Fixing the temperature ($\tau{=}0.1$) and removing the regularizer reduce accuracy to 87.37\% and 90.50\% respectively, indicating that both the adaptive temperature and its regularization contribute to the final performance. The model is also robust: accuracy remains stable across temperature settings and improves monotonically as the segmentation length grows from 5 to 30 seconds.

\begin{table}[t]
\centering
\caption{Ablation study of the proposed model on the ADFTD dataset \cite{miltiadous2023dataset}. Multi-head (5 heads) denotes multiple independently learned MLP projection heads, not frequency bands.}
\label{tab:adcn_comparison}
\setlength{\tabcolsep}{3.5pt}
\begin{tabular}{lccccc}
\toprule
\textbf{Model} & \textbf{Acc} & \textbf{F1} & \textbf{Prec.} & \textbf{Rec.} & \textbf{AUC}\\
\midrule
\textbf{Adaptive 5 band heads} & \textbf{92.37} & \textbf{92.33}  & \textbf{92.61} & \textbf{92.37} & \textbf{93.18}  \\
\midrule
w/o self-supervised learning   & 62.54    & 61.02   & 62.79 & 62.54  & 65.61 \\
constant temp. ($\tau{=}0.1$)  & 87.37    & 87.40   & 87.99 & 87.37  & 88.38   \\
Single-head                    & 89.65    & 89.62   & 90.04 & 89.65  & 91.88 \\
w/o regularization             & 90.50    & 90.47   & 91.22 & 89.89  & 91.13 \\
Multi-head (5 heads)           & 91.14    & 91.11   & 91.49 & 91.14  & 93.21 \\
\bottomrule
\end{tabular}
\end{table}

\section{Conclusion}
We presented Delta2Gamma, a multi-head SimCLR framework for contrastive EEG representation learning that uses independent CNN encoders and adaptive temperatures for each of the five frequency bands.
By computing and aggregating a per-band contrastive loss inspired by adaptive multi-head contrastive learning \cite{wang2024adaptive}, the model achieves superior representation quality and state-of-the-art AD-vs-CN classification under strict LOSO evaluation, even with limited labels. The approach is readily deployable and extensible to other biosignal-analysis domains.

\section*{Acknowledgment}
This work was supported in part by: the Institute of Information \& Communications Technology Planning \& Evaluation (IITP) grant funded by the Korean government (MSIT) under Grant No. RS-2019-II190079 for the Artificial Intelligence Graduate School Program at Korea University;
the National Research Foundation of Korea (NRF) grant funded by the Korean government (MSIT) under Grant No. RS-2025-24535409;
and the Supreme Prosecutor's Office Research Grant in 2026 (research title: Development of fake voice detection technology robust in new voice generation technology and speaker recognition).

\bibliographystyle{IEEEtran}
\bibliography{references}

\end{document}